\documentclass[10pt,twocolumn,letterpaper]{article}

\usepackage{cvpr}
\usepackage{times}
\usepackage{epsfig}
\usepackage{graphicx}
\usepackage{amsmath}
\usepackage{amssymb}
\usepackage{multirow}
\usepackage{makecell}
\usepackage{comment}
\usepackage{color}
\usepackage[caption=false]{subfig}
\usepackage{pifont}
\usepackage{float}
\usepackage{booktabs}

\usepackage[pagebackref=true,breaklinks=true,letterpaper=true,colorlinks,bookmarks=false]{hyperref}

\cvprfinalcopy

\def\cvprPaperID{****}

\begin{document}

%%%%%%%%% TITLE
\title{Dense RepPoints: Representing Visual Objects with Dense Point Sets}

\author{Ze Yang$^{1\dag}$\thanks{Equal contribution. $^\dag$This work was done when Ze Yang, Yinghao Xu and Han Xue were interns at Microsoft Research Asia.}\quad Yinghao Xu$^{2,3\dag*}$ \quad Han Xue$^{4\dag*}$ \quad Zheng Zhang$^{6}$ \\ Raquel Urtasun$^{5}$ \quad Liwei Wang$^1$ \quad Stephen Lin$^6$ \quad Han Hu$^6$ \vspace{8pt}\\
	$^1$Peking University \quad
    $^2$The Chinese University of Hong Kong  \quad
    $^3$Zhejiang University  \quad \\
    $^4$Shanghai Jiao Tong University
    $^5$University of Toronto \quad
    $^6$Microsoft Research Asia \\
    {\tt\small yangze@pku.edu.cn \quad     justimyhxu@gmail.com \quad xiaoxiaoxh@sjtu.edu.cn} \\
	{\tt\small urtasun@cs.toronto.edu \quad
	wanglw@cis.pku.edu.cn \quad
	\{zhez,stevelin,hanhu\}@microsoft.com} \\
}

\maketitle

%%%%%%%%% ABSTRACT

\begin{abstract}
  We present a new object representation, called Dense RepPoints, that utilizes a large set of points to describe an object at multiple levels, including both box level and pixel level. Techniques are proposed to efficiently process these dense points, maintaining near-constant complexity with increasing point numbers. Dense RepPoints is shown to represent and learn object segments well, with the use of a novel distance transform sampling method combined with set-to-set supervision. The distance transform sampling combines the strengths of contour and grid representations, leading to performance that surpasses counterparts based on contours or grids.
Code is available at \url{https://github.com/justimyhxu/Dense-RepPoints}.

\end{abstract}

\vspace{-10pt}

\section{Introduction}

Representation matters. While significant advances in visual understanding algorithms have been witnessed in recent years, they all rely on proper representation of visual elements for convenient and effective processing. For example, a single image feature, a rectangular box, and a mask are usually adopted to represent input for recognition tasks of different granularity, i.e.~image classification~\cite{AlexNet,ResNet,VGG}, object detection~\cite{girshick2015fast,ren2015faster,RetinaNet} and pixel-level segmentation~\cite{Mask-rcnn,FCIS,TensorMask}, respectively. In addition, the representation at one level of granularity may help the recognition task at another granularity, e.g.~an additional mask representation may aid the learning of a coarser recognition task such as object detection~\cite{Mask-rcnn}. We thus consider the question of whether a unified representation for recognition tasks can be devised over various levels of granularity.

Recently, RepPoints~\cite{RepPoints} was proposed to represent an object by a small set of adaptive points, simultaneously providing a geometric and semantic description of an object. It demonstrates good performance for the coarse localization task of object detection, and also shows potential to conform to more sophisticated object structures such as semantic keypoints. However, the small number of points (9 by default) limits its ability to reveal more detailed structure of an object, such as pixel-level instance segmentation. In addition, the supervision for recognition and coarse localization also may hinder learning of more fine-grained geometric descriptions.

\begin{figure}[tb]
\begin{center}
\includegraphics[width=0.48\textwidth]{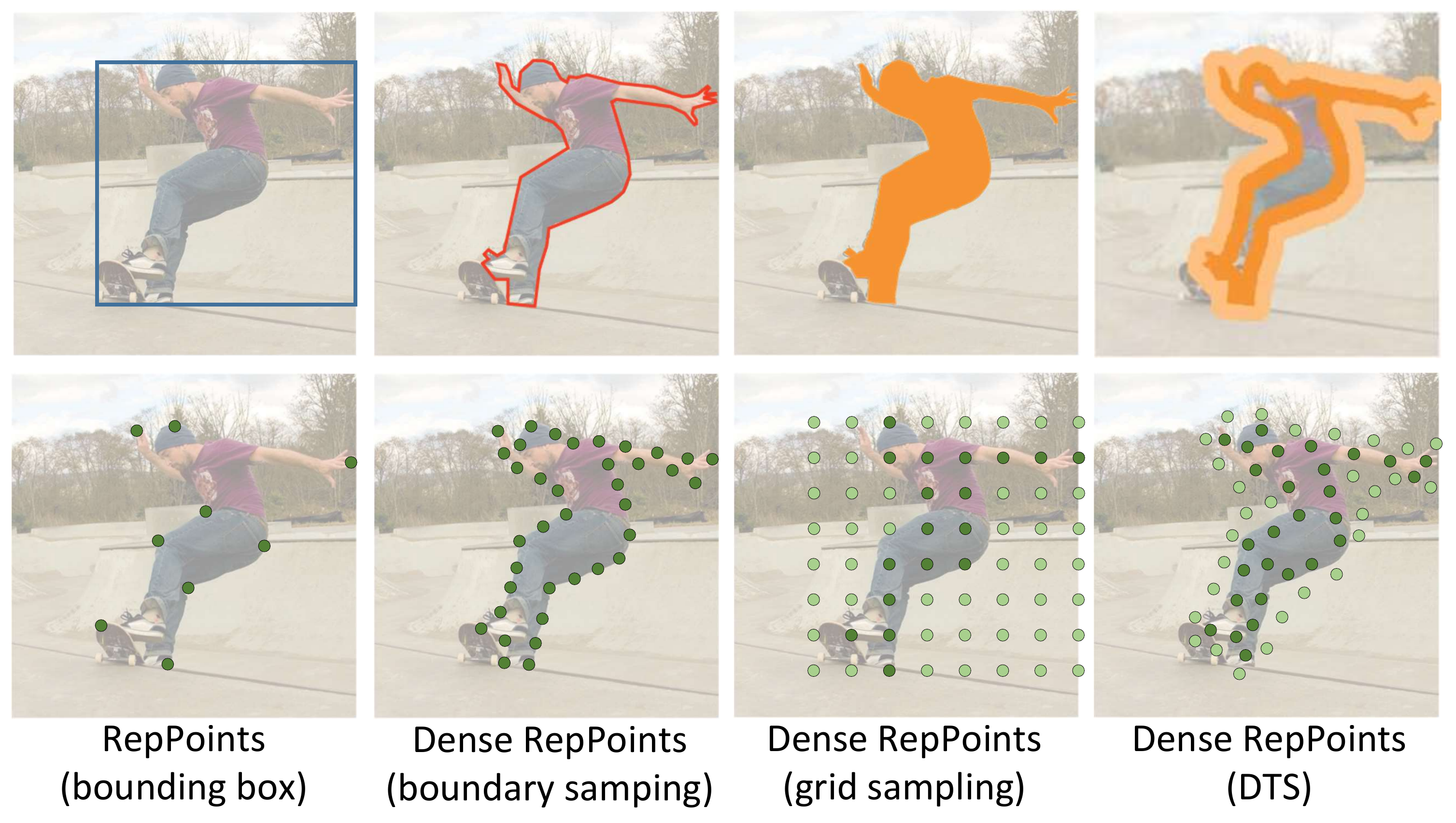}
\caption{
\textbf{A visual object modeled in different geometric forms (top row from left to right)}: bounding box, boundary sampling (contour), grid sampling (binary mask), distance transform sampling (binary boundary mask). These various object forms are modeled in this work by a unified representation consisting of a dense point set, called {\em Dense RepPoints} (bottom row).
}
\label{fig::representation}
\end{center}
\vspace{-25pt}
\end{figure}

This paper presents \emph{Dense RepPoints}, which utilizes a large number of points along with optional attributes to represent objects in detail, e.g.~for instance segmentation. Because of its high representation flexibility, \emph{Dense RepPoints} can effectively model common object segment descriptors, including contours (polygon)~\cite{kass1988snakes, MSCOCO,PolarMask,kass1988snakes} and grid masks~\cite{Mask-rcnn,TensorMask}, as illustrated in columns 2 and 3 of Figure~\ref{fig::representation}. \emph{Dense RepPoints} can also model a \emph{binary boundary mask}, a new geometric descriptor for object segments that combines the description efficiency of contours and the reduced dependence on exact point localization of grid masks, as illustrated in column 4 of Figure~\ref{fig::representation}.

To learn and represent binary boundary masks by \emph{Dense RepPoints}, three techniques are proposed. The first is a distance transform sampling (DTS) method, which converts a ground-truth boundary mask into a point set by probabilistic sampling based on the distance transform map of the object contour. With this conversion, the \emph{Dense RepPoints} prediction and ground truth are both point sets and are thus comparable. The second is a set-to-set supervision loss, in contrast to the commonly used point-to-point supervision loss, e.g.~\cite{PolarMask, peng2020deepsnake}. The set-to-set supervision loss avoids assigning exact geometric meaning for every point, which is usually difficult and  semantically inaccurate for instance segmentation but is required by point-to-point methods. The third is a novel conversion method from the learnt non-grid \emph{Dense RepPoints} to an instance mask of any resolution, based on Delaunay triangulation. 

With these three novel techniques, \emph{Dense RepPoints} are learnt to well represent the binary boundary map of objects. It also yields better performance than methods based on a contour or grid mask representation. The method achieves 39.1 mask mAP and 45.6 box mAP on the COCO test-dev set using a ResNet-101 backbone network.

In addition to greater representation ability and better accuracy, Dense RepPoints can also be efficiently processed with our proposed techniques. The complexity of vanilla RepPoints increases linearly with the number of points, making it impractical for large point sets, e.g. hundreds of points. To resolve this issue, we propose two techniques, \textit{group pooling} and \textit{shared offset / attribute field}, for object classification and offset / attribute prediction, respectively. These techniques enable near-constant complexity with increasing numbers of points, while maintaining the same accuracy. 

The contributions of this work are summarized as follows:
\vspace{-2pt}
\begin{itemize}
\item We propose a new object representation, called \emph{Dense RepPoints}, that models objects by a large number of adaptive points. The new representation shows great flexibility in representing detailed geometric structure of objects. It also provides a unified object representation over different levels of granularity, such as at the box level and pixel level. This allows for coarse detection tasks to benefit from finer segment annotations as well as enable instance segmentation, in contrast to training through separate branches built on top of base features as popularized in \cite{MNC,Mask-rcnn}.
 \item We adapt the general \emph{Dense RepPoints} representation model to the instance segmentation problem, where three novel techniques of \emph{distance transform sampling} (DTS), \emph{set-to-set supervision loss} and \emph{Delaunay triangulation based conversion} are proposed. \emph{Dense RepPoints} is found to be superior to previous methods built on a contour or grid mask representation.
\item We propose two techniques, of \emph{group pooling} and \emph{shared offset / attribute fields}, to efficiently process the large point set of \emph{Dense RepPoints}, yielding near constant complexity with similar accuracy.
\end{itemize}

\section{Related Work}

\paragraph{Bounding box representation.}
Most existing high-level object recognition benchmarks \cite{PascalVOC,MSCOCO,OpenImagesV4} employ bounding box annotations for object detection. The current top-performing two-stage object detectors \cite{girshick2014rich,girshick2015fast,ren2015faster,dai2016r} use bounding boxes as anchors, proposals and final predictions throughout their pipelines. Some early works have proposed to use rotated boxes \cite{huang2007high} to improve upon axis-aligned boxes, but the representation remains in a rectangular form. For other high-level recognition tasks such as instance segmentation and human pose estimation, the intermediate proposals in top-down solutions~\cite{MNC,Mask-rcnn} are all based on bounding boxes. However, the bounding box is a coarse geometric representation which only encodes a rough spatial extent of an object.

\vspace{-5pt}

\paragraph{Non-box object representations.}
For instance segmentation, the annotation for objects is either as a binary mask \cite{PascalVOC} or as a set of polygons \cite{MSCOCO}. While most current top-performing approaches \cite{InstFCN,Mask-rcnn,chen2018masklab} use a binary mask as final predictions, recent approaches also exploit contours for efficient interactive annotation \cite{PolygonRNN,PolygonRNN++,CurveGCN} and segmentation \cite{DARNet,PolarMask}.  This contour representation, which was popular earlier in computer vision \cite{kass1988snakes,chan2001active,srinivasan2010many,toshev2012shape,wang2012fan}, is believed to be more compatible with the semantic concepts of objects \cite{palmer1999vision,toshev2012shape}. Some works also use edges and superpixels \cite{yang2016object,InstanceCut} as object representations. Our proposed \emph{Dense RepPoints} has the versatility to model objects in several of these non-box forms, providing a more generalized representation. 

\begin{figure*}[ht]
\centering
\includegraphics[width = 0.98\linewidth]{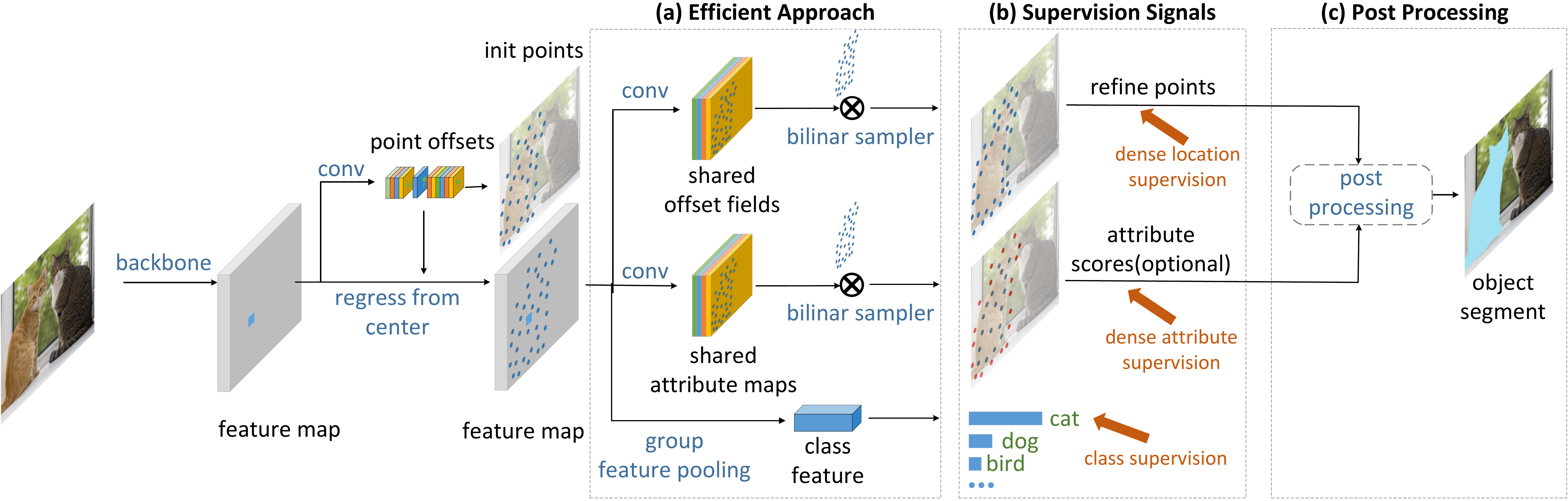}
\vspace{-3pt}
\caption{Overview of \textit{Dense RepPoints}. First, the initial representative points are generated by regressing from the center point as in RepPoints~\cite{RepPoints}. Then, these initial points are refined by the proposed efficient approaches to obtain refined, attributed representative points. Finally, post-processing is applied to generate the instance segment.}
\label{fig::pipeline}
\vspace{-10pt}
\end{figure*}

\vspace{-5pt}

\paragraph{Point set representation.}
There is much research focused on representing point clouds in 3D space  \cite{PointNet,PointNet++}. A direct instantiation of ordered point sets in 2D perception is 2D pose \cite{CPM,cao2017realtime,densepose}, which directly addresses the semantic correspondence problem. Recently, there has been increasing interest in the field of object detection on using specific point locations, including corner points \cite{CornerNet}, extreme points \cite{ExtremeNet}, and the center point \cite{CenterNet, duan2019centernet}. These point representations are actually designed to recover a bounding box, which is coarse and lacks semantic information. RepPoints \cite{RepPoints} proposes a learnable point set representation trained from localization and recognition feedback. However, it uses only a small number ($n=9$) of points to represent objects, limiting its ability to represent finer geometry. In this work, we extend RepPoints \cite{RepPoints} to a denser and finer geometric representation, enabling usage of dense supervision and taking a step towards dense semantic geometric representation.

\section{Methodology}

\label{sec::dense_reppoints}
In this section, we first review \textit{RepPoints}~\cite{RepPoints} for object detection in Sec.~\ref{sec::review_reppoint}. Then, we introduce \textit{Dense RepPoints} in Sec.~\ref{sec::dense_reppoint} for strengthening the representation ability of \textit{RepPoints} from object detection to fine-grained geometric localization and recognition tasks, such as extracting an instance mask, by associating an attribute vector with each representative point. In addition, these fine-grained tasks usually require higher resolution and many more representative points than object detection, which makes the computational complexity of vanilla \textit{RepPoints} infeasible. We discuss how to reduce the computational complexity of vanilla \textit{RepPoints} in Sec.~\ref{sec::efficient} for representing an instance mask. In Sec.~\ref{sec::mask_representations}, we describe how to use \textit{Dense RepPoints} to model instance masks with different sampling strategies, and then design appropriate supervision signals in Sec.~\ref{sec::mask_supervision}. Since representative points are usually sparse and non-grid while an instance segment is dense and grid-aligned, we discuss how to transform representative points into an instance segment in Sec.~\ref{sec::mask_transform}. An overview of our method is exhibited in Fig.~\ref{fig::pipeline}.

\subsection{Review of RepPoints for object detection}
\label{sec::review_reppoint}
We first review how \textit{RepPoints}~\cite{RepPoints} detects objects. A set of adaptive representative points $\mathcal{R}$ is used to represent an object in RepPoints:
\vspace{-3pt}
\begin{align}
\label{eq::RepPoints}
    \mathcal{R} = \{p_i\}_{i=1}^{n}
\end{align}
where $p_i=(x_i+\Delta x_i$, $y_i + \Delta y_i)$ is the $i$-th representative point, $x_i$ and $y_i$ denote an initialized location, $\Delta x_i$ and $\Delta y_i$ are learnable offsets, and $n$ is the number of points. The feature of a point $\mathcal{F}(p)$ is extracted from the feature map $\mathcal{F}$ through bilinear interpolation, and the feature of a point set $\mathcal{F}(\mathcal{R})$ is defined as the concatenation of all representative points of $\mathcal{R}$:
\vspace{-3pt}
\begin{align}
    \mathcal{F}(\mathcal{R}) = \text{concat}(\mathcal{F}(p_1), ..., \mathcal{F}(p_n))
\end{align}
which is used to recognized the class of the point set. The bounding box of a point set can be obtained by a conversion function. In the training phase, explicit supervision and annotation for representative points is not required. Instead, representative points are driven to move to appropriate locations by the box classification loss and box localization loss:
\vspace{-3pt}
\begin{align}
\label{eq::obj_det}
L_\text{det} =L^{b}_\text{cls}+L^{b}_\text{loc}
\end{align}
as both bilinear interpolation used in feature extraction and the conversion function used in bounding box transformation are differentiable with respect to the point locations. These representative points are suitable for representing the object category and accurate position at the same time.

\subsection{Dense RepPoints}
\label{sec::dense_reppoint}
In vanilla \textit{RepPoints}, the number of representative points is relatively small $(n=9)$. It is sufficient for object detection, since the category and bounding box of an object can be determined with few points. Different from object detection, fine-grained geometric localization tasks such as instance segmentation usually provide pixel-level annotations that require precise estimation.
%, and the goal is changed from predicting the attribution of a point set to each point. 
Therefore, the representation capacity of a small number of points is insufficient, and a significantly larger set of points is necessary together with an attribute vector associated with each representative point:
\vspace{-3pt}
\begin{align}
\label{eq::DenseRepPoints}
    \mathcal{R} = \{(x_i+\Delta x_i, y_i+\Delta y_i, \mathbf a_i)\}_{i=1}^{n},
\end{align}
where $\mathbf a_i$ is the attribute vector associated with the $i$-th point. 

In instance segmentation, the attribute can be a scalar, defined as the foreground score of each point. In addition to the box-level classification and localization terms, $L^{b}_\text{cls}$ and $L^{b}_\text{loc}$, we introduce a point-level classification loss $L^{p}_\text{cls}$ and a point-level localization loss $L^{p}_\text{loc}$. The objective function of Eq.~\ref{eq::obj_det} becomes:
\vspace{-3pt}
\begin{align}
\label{eq::obj_seg}
L = \underbrace{L^{b}_\text{cls}+L^{b}_\text{loc}}_{L_\text{det}}+\underbrace{L^{p}_\text{cls}+L^{p}_\text{loc}}_{L_\text{mask}}
\end{align}
where $L^{p}_{cls}$ is responsible for predicting the point foreground score and $L^{p}_{loc}$ is for learning point localization. This new representation is named \textit{Dense RepPoints}. 

\subsection{Efficient computation}
\label{sec::efficient}
Intuitively, denser points will improve the capacity of the representation. However, the feature of an object in vanilla \textit{RepPoints} is formed by concatenating the features of all points, so the FLOPs will rapidly increase as the number of points increases. Therefore, directly using a large number of points in \textit{RepPoints} is impractical. To address this issue, we introduce group pooling and shared offset fields to reduce the computational complexity, thereby significantly reducing the extra FLOPs while maintaining performance. In addition, we further introduce a shared attribute map to efficiently predict whether a point is in the foreground. 

\begin{figure}
\centering
\includegraphics[width=0.45\textwidth]{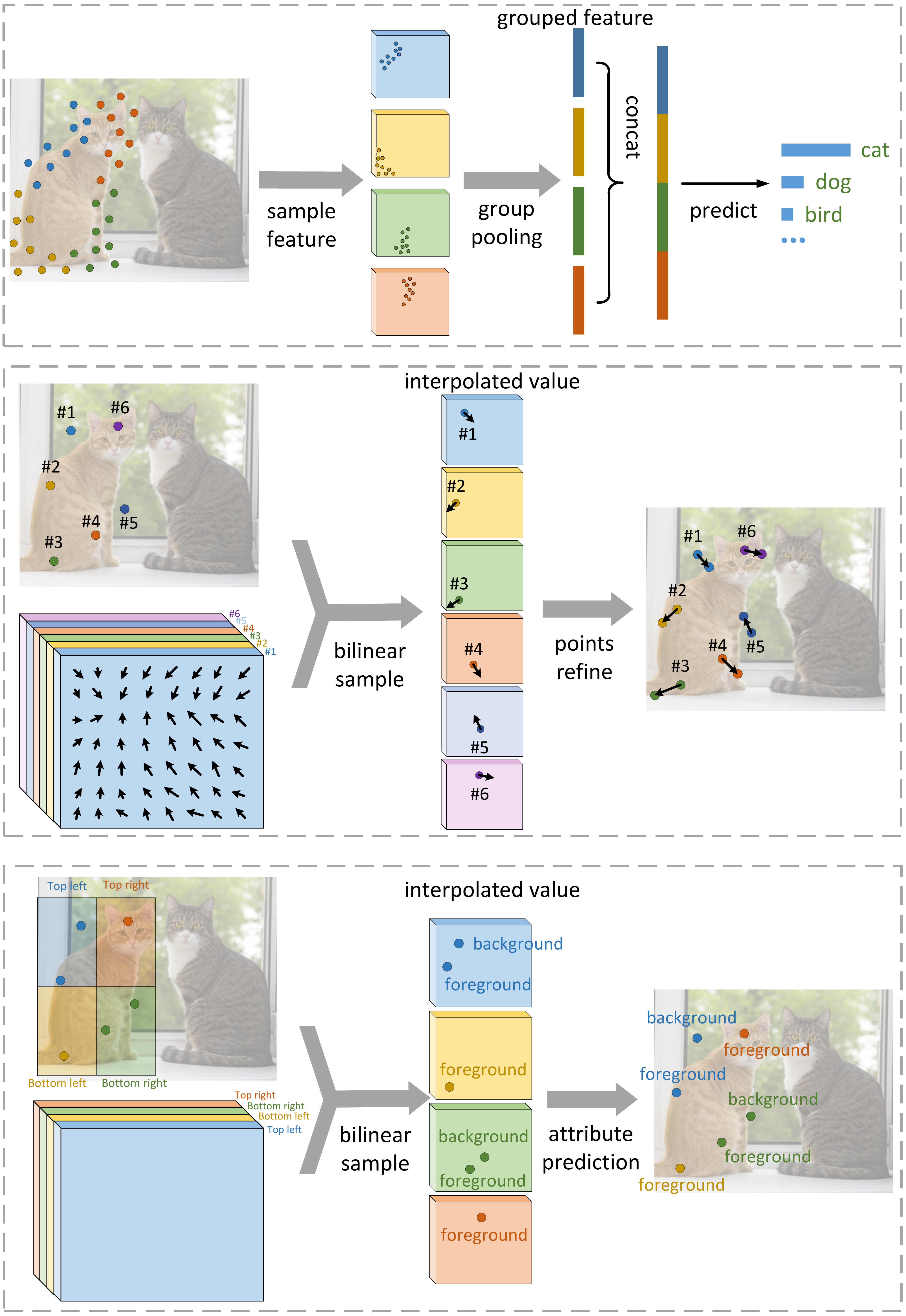}
\caption{Illustration of efficient feature extraction for \textit{Dense RepPoints}. \textbf{Top:} group pooling operation. \textbf{Middle:} shared offset fields for each point index. \textbf{Bottom:} shared attribute maps for each relative position.}
\label{fig::efficient}
\vspace{-10pt}
\end{figure}

\vspace{-10pt}
\paragraph{Group pooling.}
Group pooling is designed to effectively extract object features and is used in the box classification branch (see Figure \ref{fig::efficient} top). Given $n$ representative points, we equally divide the points into $k$ groups, with each group having $n/k$ points (if $k$ is not divisible by $n$, the last group will have fewer points than the others to ensure a total of $n$ points). Then, we aggregate the feature of each point within a group by max-pooling to extract a group feature. Finally, a $1\times 1$ convolution is computed over the concatenated group features from all groups. In this way, the object features are represent by groups instead of points, reducing the computational complexity from $O(n)$ to $O(k)$. We empirically find that the number of groups do not need to be increased when the points become denser, thus the computational complexity is not affected by using a larger set of points. In our implementation, we set $k$ to 9 by default, which works relatively well for classification. 

\vspace{-10pt}
\paragraph{Shared offset fields.}
The computational complexity of predicting the offsets for the points is $O(n^2)$ in \textit{RepPoints}, making the dense point set representation unsuitable for real applications. Unlike in the classification branch, we need the information of individual points for point location refinement. Hence, we cannot directly apply the grouped features used in classification. Instead, we empirically find that local point features provide enough information for point refinement, in the same spirit as Curve-GCN~\cite{CurveGCN} which uses local features for contour refinement. To share feature computation among points, we propose to first compute $n$ shared offset field maps based on the image feature map. And then for the $i$-th representative point, its position is directly predicted via bilinear interpolation at the corresponding location of the $i$-th offset field (see Figure \ref{fig::efficient} middle). This reduces the computational complexity of the regression from $O(n^2)$ to $O(n)$. By using group pooling and shared offset fields, even if a large number of points are used, the added FLOPs is still very small compared to that of the entire network (see Sec.~\ref{sec::ablation}).

\vspace{-10pt}
\paragraph{Shared attribute map.}
Predicting the foreground score of each point can be implemented in manner similar to \textit{shared offset fields} by using a shared position-sensitive attribute map, first introduced in R-FCN~\cite{dai2016r}. In the position-sensitive attribute map, each channel has an explicit positional meaning. Therefore, the foreground score of each representative point can be interpolated on the channel corresponding to its location (see Figure~\ref{fig::efficient} bottom). 

\subsection{Different sampling strategies}
\label{sec::mask_representations}
How to represent object segments effectively is a core problem in visual perception.
Contours and binary masks are two typical representations widely used in previous works~\cite{Mask-rcnn,TensorMask,PolarMask,peng2020deepsnake}. In \textit{Dense RepPoints}, these representations can be simulated by different sampling strategies: a binary mask by uniformly sampling grid points over the bounding box of an object, and a contour as all sampling points along the object boundary. We call these two sampling strategies \textit{grid sampling} and \textit{boundary sampling}, respectively, and discuss them in this section. In addition, we introduce a new sampling strategy, named \textit{distance transform sampling}, which combines the advantages of both grid sampling and boundary sampling.

\vspace{-10pt}
\paragraph{Boundary sampling (Contour).}
An instance segment can be defined as the inner region of a closed object contour. Contour points is a compact object description because of its 1-D nature (defined by a sequence of points). In our method, the contour representation can be simulated through supervising the offsets of representative points along the object boundary, with the score of points set to 1 by default.

\vspace{-10pt}
\paragraph{Grid sampling (Binary Mask).}
A binary mask can be represented as a set of uniformly sampled grid points over the bounding box of an object, and each sampled point has a binary score to represent its category, i.e. foreground or background. This sampling strategy (representation) is widely used, such as in Mask R-CNN~\cite{Mask-rcnn} and Tensor Mask~\cite{TensorMask}. In \textit{Dense RepPoints}, grid sampling can be implemented by constraining the offsets of representative points as:
\begin{align}
\Delta x_{i}= \alpha (\frac{i}{\sqrt{n}} - 0.5), i \in \{1...n\} \\
\Delta y_{i} = \beta (\frac{i}{\sqrt{n}} - 0.5), i \in \{1...n\}
\end{align}
where $n$ is the number of sampling points, and $\alpha$ and $\beta$ are two learnable parameters. 

\vspace{-10pt}
\paragraph{Distance transform sampling (Binary Boundary Mask).}
Boundary sampling and grid sampling both have their advantages and applications. In general, boundary sampling (contour) is more compact for object segment description, and grid sampling is easier for learning, mainly because its additional attribute (foreground score) avoids the need for precise point localization. To take advantage of both sampling strategies, we introduce a new sampling method called distance transform sampling. In this sampling strategy, points near the object boundary are sampled more and other regions are sampled less. During the training phase, the ground truth is sampled according to distance from the object boundary:
\begin{align}
    \label{eq::prob}
    \mathcal{P}(p) & = \frac{g(D(p))}{\sum_q g(D(q))}\\
    \label{eq::dist}
    \mathcal{D}(p) & =  \frac{\mathop{\min}_{e\in\mathcal{E}}\lVert p-e \rVert_2}{\sqrt{\mathop{\max}_{e,e'\in\mathcal{E}}\left| e_x-e'_x \right | \cdot \mathop{\max}_{e,e'\in\mathcal{E}}\left| e_y-e'_y \right|}}
\end{align}
% _{e\in\mathcal{E}}
where $P(p)$ is the sampling probability of point $p$, $D(p)$ is the normalized distance from the object boundary of point $p$, $\mathcal{E}$ is the boundary point set, and $g$ is a decreasing function. In our work, we use the Heaviside step function for $g$:
\begin{align}
g(x) = \begin{cases}
1& x\leq \delta\\
0& x> \delta
\end{cases}
\end{align}
Here, we use $\delta=0.04$ by default. Intuitively, points with a distance less than $\delta$ (close to the contour) have a uniform sampling probability, and points with a distance greater than $\delta$ (away from the contour) are not sampled.

\subsection{Sampling supervision}
\label{sec::mask_supervision}
The point classification loss $L^p_\text{cls}$ and the point localization loss $L^p_\text{loc}$ are used to supervise the different segment representations during training. In our method, $L^p_\text{cls}$ is defined as a standard cross entropy loss function with softmax activation, where a point located in the foreground is labeled as positive and otherwise its label is negative. %Localization supervision has two methods: the point-to-point supervision and set-to-set supervision. 

For localization supervision, a point-to-point approach could be taken, where each ground truth point is assigned an exact geometric meaning, e.g. using the polar assignment method in PolarMask~\cite{PolarMask}.
%or the organized assignment method in v1 of this paper. 
Each ground truth with exact geometric meaning also corresponds to a fixed indexed representative point in \emph{Dense RepPoints}, and the L2 distance is used as the point localization loss $L_{loc}^p$:
\vspace{-3pt}
\begin{align}
L_{point}(\mathcal{R}, \mathcal{R}')=\frac{1}{n}\sum\limits_{k=1}^n \left \lVert (x_i, y_i) - (x_i', y_i') \right \rVert_2
\end{align}
where $(x_i, y_i)\in \mathcal{R}$ and $(x_i', y_i')\in \mathcal{R'}$ represent the point in the predicted point set and ground-truth point set, respectively.

However, assigning exact geometric meaning to each point is difficult and may be semantically inaccurate for instance segmentation.
Therefore, we propose set-to-set supervision, rather than supervise each individual point. The point localization loss is measured by \textit{Chamfer distance} \cite{fan2017point,rubner2000earth} between the supervision point set and the learned point set:
\vspace{-3pt}
\begin{align}
\nonumber L_{set}(\mathcal{R}, \mathcal{R}') & =\frac{1}{2n} \sum\limits_{i=1}^n\min\limits_j{\left\lVert (x_i, y_i) - (x_
j', y_j')\right\rVert_2} \\
&+\frac{1}{2n}\sum\limits_{j=1}^n\min\limits_i\left\lVert (x_i, y_i) - (x_
j', y_j')\right\rVert_2
\end{align}
where $(x_i, y_i)\in \mathcal{R}$ and $(x_j', y_j')\in \mathcal{R'}$. We evaluate these two forms of supervision in Section~\ref{sec::ablation}. 

\subsection{Representative Points to Object Segment}
\label{sec::mask_transform}
\emph{Dense RepPoints} represents an object segment in a sparse and non-grid form, and thus an extra post-processing step is required to transform the non-grid points into a binary mask. In this section, we propose two approaches, Concave Hull~\cite{moreira2007concave} and Triangulation, for this purpose. 

\begin{figure}
\centering
\includegraphics[width=0.48\textwidth]{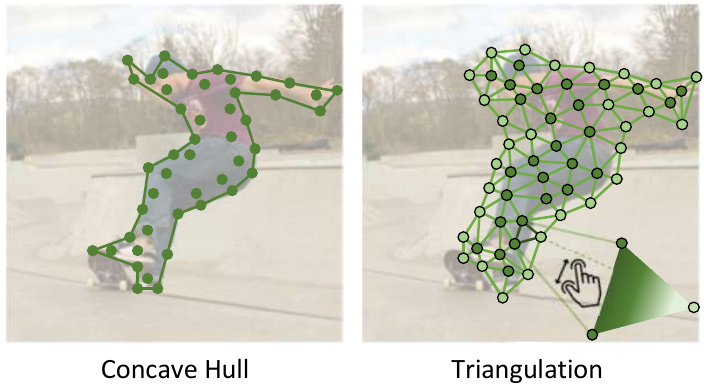}
\caption{\textbf{Post-Processing}. Generating image segments by Concave Hull and Triangulation.}
\label{fig::inference}
\vspace{-10pt}
\end{figure}

\vspace{-10pt}
\paragraph{Concave Hull.}
An instance mask can be defined as a concave hull of a set of foreground points (see Figure~\ref{fig::inference} left), which is used by many contour-based methods. In \textit{Dense RepPoints}, boundary sampling naturally uses this post-processing. We first use a threshold to binarize the predicted points by their foreground scores, and then compute their concave hull to obtain the binary mask. In our approach, we empirically set a threshold of $0.5$ by default.

\vspace{-10pt}
\paragraph{Triangulation.}
Triangulation is commonly used in computer graphics to obtain a mesh from a point set representation, and we introduce it to this problem to generate an object segment. Specifically, we first apply Delaunay triangulation to partition the space into triangles with vertices defined by the learned point set. Then, each pixel in the space will fall inside a triangle and its point score is obtained by linearly interpolating from the triangle vertices in Barycentric coordinates (Figure~\ref{fig::inference} right). Finally, a threshold is used to binarize the interpolated score map to obtain the binary mask.

\section{Experiments}
\begin{figure*}[ht]
\centering
\includegraphics[width=0.93\textwidth]{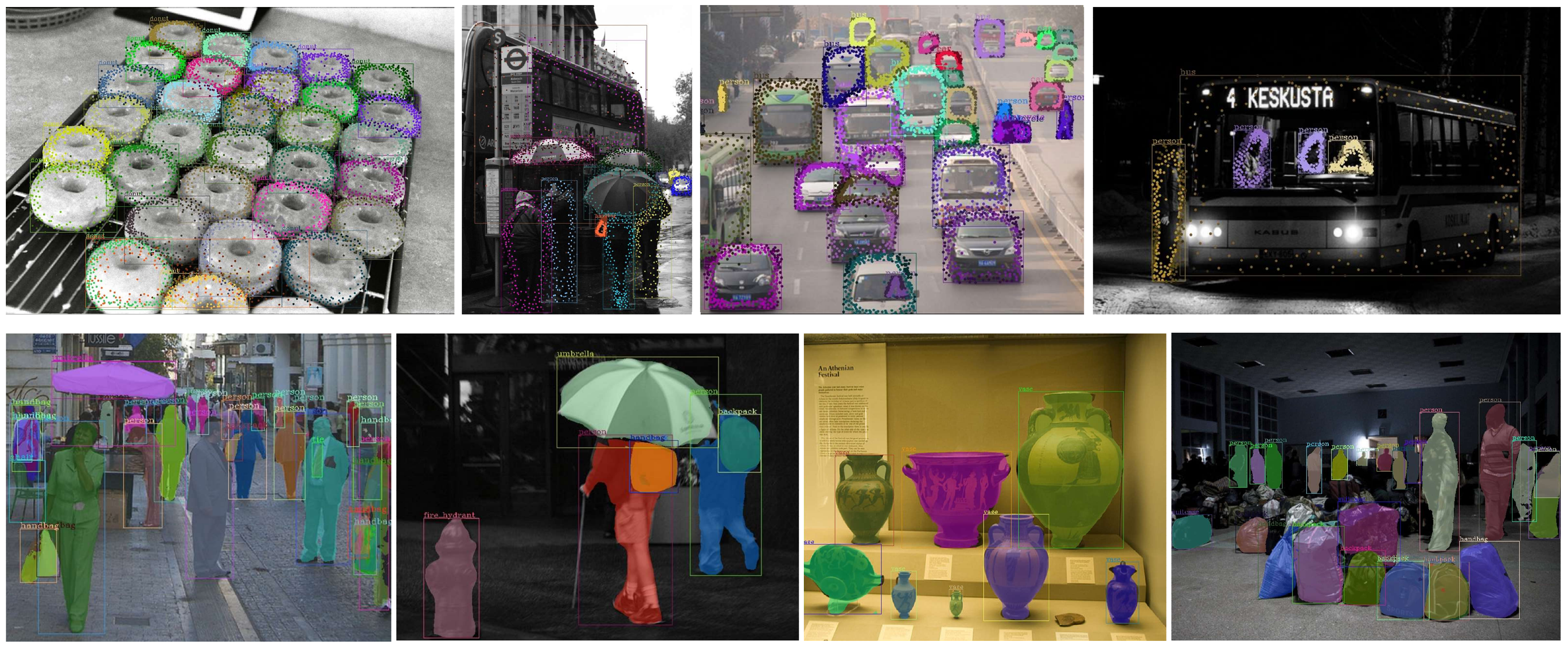}
\caption{Visualization of points and instance masks by DTS. 
\textbf{Top}:The learned points (225 points) is mainly distributed around the mask boundary.
\textbf{Bottom}: The foreground masks generated by triangulation post-processing on COCO \texttt{test-dev} images with ResNet-50 backbone under '3x' training schedule.}
\label{fig::vis}
\vspace{-12pt}
\end{figure*}

\subsection{Datasets}
We present experimental results for instance segmentation and object detection on the COCO2017 benchmark~\cite{MSCOCO}, which contains 118k images for training, 5k images for validation (\texttt{val}) and 20k images for testing (\texttt{test-dev}). The standard mean average precision (mAP) is used to measure accuracy. We conduct an ablation study on the validation set, and compare with other state-of-the-art methods on the test-dev set.

\subsection{Implementation Details}
We follow the training settings of \textit{RepPoints}~\cite{RepPoints}. Horizontal image flipping augmentation, group normalization~\cite{GN} and focal loss~\cite{RetinaNet} are used during training. If not specified, ResNet-50~\cite{ResNet} is used as the default backbone in the ablation study, and weights are initialized from the ImageNet~\cite{deng2009imagenet} pretrained model. Distance transform sampling with set-to-set supervision is used as the default training setting, and triangulation is chosen as the default post-processing. For predicting attribute scores, we follow SOLO~\cite{wang2019solo} by using seven $3\times3$ convs in the attribute score head. 

The models are trained on $8$ GPUs with $2$ images per GPU for $12$ epochs ($1\times$ settings). In SGD training, the learning rate is initialized to $0.01$ and then divided by $10$ at epochs $8$ and $11$. The weight decay and momentum parameters are set to $10^{-4}$ and $0.9$, respectively. In inference, we follow SOLO~\cite{wang2019solo} to refine the classification score by using the mask prediction, and we use NMS with IoU threshold of $0.5$, following RetinaNet~\cite{RetinaNet}.

\begin{table}[t]
    \caption{Validating the proposed components for greater efficiency. With group pooling (GP) and shared offset fields (SOF), the mAP constantly improve as the number of points increases, while the FLOPS is nearly unaffected.}
    \vspace{-10pt}
    \begin{center}
    \begin{tabular}{c|c|c|c|c|c|c}
    \hline
    \multirow{2}{*}{n}   & \multicolumn{3}{c|}{G FLOPS} & \multicolumn{3}{c}{mAP} \\ \cline{2-7}
      & Base    & + GP    & +SOF    & Base  & + GP  & + SOF   \\ 
    \hline \hline 
    9  & 211.04  & 208.03   & 202.05   & 38.1  & 37.9   & 37.9  \\ 
    \hline
    25 & 255.14  & 237.80   & 205.93   & 37.7  & 37.8   & 37.7  \\ 
    \hline
    49 & 321.28  & 278.86   & 209.18   & 37.7  & 37.6   & 37.5  \\ 
    \hline
    81 & 409.46  & 331.03   & 212.60   & 37.5  & 37.5   & 37.5  \\ 
    \hline
    \end{tabular}
    \end{center}
    \label{tab::efficient}
    \vspace{-20pt}
\end{table}

\subsection{Ablation Study}
\label{sec::ablation}
\paragraph{Components for greater efficiency}
We validate group pooling (GP) and shared offset fields (SOF) by adding them to vanilla \textit{RepPoints}~\cite{RepPoints} and evaluating the performance on object detection. Results are shown in Table~\ref{tab::efficient}. We present the results under different numbers of points: $n=9,25,49,81$. By using group pooling, FLOPs significantly decreases with increasing number of points compared to vanilla \textit{RepPoints} with similar mAP. By introducing shared offset fields, while mAP is not affected, FLOPs is further reduced and nearly constant with respect to $n$. Specifically, for $n=81$, our efficient approach saves $197$G FLOPs in total. This demonstrates the effectiveness of our efficient approach representation and makes the use of more representative points in instance segmentation possible.

\begin{table}[t]
    \caption{Comparison of different mask representations.}
    \vspace{+5pt}
    \begin{center}
    \begin{tabular}{c|c|c|c|c|c}
    \hline
    number of points & $9$ & $25$  & $81$ & $225$ & $729$ \\
    \hline
    \hline
    Contour & \textbf{19.7} & 23.9 & 26.0 & 25.2 & 24.1 \\
    \hline
    Grid points & 5.0 & 17.6 & 29.7 & 31.6 & 32.8\\
    \hline
    DTS & 13.9 & \textbf{24.5} & \textbf{31.5} & \textbf{32.8} & \textbf{33.8}\\
    \hline
    \end{tabular}
    \end{center}
    \label{tab::non_bbox}
    \vspace{-15pt}
\end{table}

\vspace{-10pt}
\paragraph{Different sampling strategies.}
We compare different strategies for sampling object points. Since different sampling strategies perform differently under different post-processing, we compare them with the their best-performing post-processing method for fair comparison. Therefore, we use triangulation (Figure~\ref{fig::inference} right) for distance transform sampling, bilinear interpolation (\texttt{imresize}) for grid sampling, and concave hull (Figure~\ref{fig::inference} left) for boundary sampling. Please see the Appendix for more details on the post-processing. Results are shown in Table~\ref{tab::non_bbox}. Boundary sampling has the best performance with few points. When $n=9$, boundary sampling obtains 19.7 mAP, and grid sampling has only 5.0 mAP. Distance transform sampling has 13.9 mAP, which lies in the middle. The reason is that boundary sampling only samples points on the boundary, which is the most efficient way to represent object masks, so relatively good performance can be achieved with fewer points. Both grid sampling and distance transform sampling need to sample non-boundary points, so their efficiency is lower than boundary sampling, but distance transform sampling samples more points around the boundary than in other regions, thus it performs much better than grid sampling. 

When using more points, grid sampling and distance transform sampling perform better than boundary sampling. For $n=729$, grid sampling and distance transform sampling achieve 32.8 mAP and 33.8 mAP, respectively, while boundary sampling only obtains 24.1 mAP. This is due to the limited representation capability of boundary sampling since it only takes boundary points into consideration. In addition, distance transform sampling outperforms grid sampling in all cases, which indicates that distance transform sampling is more efficient than grid sampling while maintaining the same representation capability.

\begin{table}[ht]
    \caption{Comparison of triangulation and concave hull.}
    \vspace{+5pt}
    \begin{center}
    \begin{tabular}{c|c|c|c|c|c }
        \hline
         \# of points & 9 & 25 & 49 & 81 & 225  \\
        \hline
        \hline
        Concave-Hull & 9.7 & 21.0 & 21.3 &  20.6  & 23.4\\
        \hline
        Triangulation & 13.9 & 24.5 & 29.6 & 31.5 & \textbf{32.8} \\
        \hline
    \end{tabular}
    \end{center}
    \label{tab::post-processing}
    \vspace{-20pt}
\end{table}

\paragraph{Concave Hull vs. Triangulation.}
Concave hull and triangulation both can transform a point set to a binary mask. Here, we compare them using distance transform sampling. Results are shown in Table~\ref{tab::post-processing}. Triangulation outperforms concave hull consistently with different numbers of points, indicating that triangulation is more suitable for distance transform sampling. It is noted that concave hull with distance transform sampling is worse than contour, because distance transform sampling does not strictly sample on the boundary but usually samples points near the boundary. Besides, it also samples points farther from the boundary.

\begin{table}[t]
    \caption{Comparison of point-to-point and set-to-set supervision.}
    \vspace{-5pt}
    \begin{center}
    \begin{tabular}{c|c|c|c|c|c}
        \hline 
         number of points & 9 & 25 & 81 & 225 & 729 \\
        \hline 
        \hline
         point-to-point     & 10.7  & 20.7  & 27.8 & 31.3 & 32.6 \\
        \hline
         set-to-set  & 13.9 & 24.5  & 31.5 & 32.8 & \textbf{33.8} \\
        \hline
    \end{tabular}
    \end{center}
    \label{tab::organize}
    \vspace{-10pt}
\end{table}

\vspace{-10pt}
\paragraph{Different supervision strategies.}
Point-to-point is a common and intuitive supervision strategy and it is widely used by other methods~\cite{PolarMask,ExtremeNet,peng2020deepsnake}. However, this kind of supervision may prevent \textit{Dense RepPoints} from learning better sampling strategies, since it is restrictive and ignores the relationships among points. This motivates the proposed set-to-set supervision in Section~\ref{sec::mask_supervision}. We compare the two forms of supervision using distance transform sampling. Results are shown in Table~\ref{tab::organize}. Set-to-set supervision consistently outperforms point-to-point supervision, especially for a small number of points.

\begin{table}[t]
\caption{Results of \textit{Dense RepPoints} on different numbers of points.}
\vspace{-5pt}
\begin{center}
\begin{tabular}{c|c|c|c|c}
    \hline
     number of points &  81  & 225 & 441 & 729  \\
    \hline
    \hline
    AP & 31.5 & 32.8 & 33.3 & \textbf{33.8} \\
    \hline
    AP@50 & 54.2 & 54.2 & 54.5 & \textbf{54.8} \\
    \hline
    AP@75 & 32.7 & 34.4 & 35.2 & \textbf{35.9} \\
    \hline
\end{tabular}
\end{center}
\label{tab::point_num}
\vspace{-10pt}
\end{table}

\vspace{-10pt}
\paragraph{More representative points.}
\textit{Dense RepPoints} can take advantage of more points than vanilla \textit{RepPoint}~\cite{RepPoints}, and its computation cost does not change as the number of points increases. Table~\ref{tab::point_num} shows the performance of \textit{Dense RepPoints} on different numbers of points using distance transform sampling and triangulation inference. In general, more points bring better performance, but as the number of points increases, the improvement saturates.

\begin{table}[t]
    \caption{Effects of dense supervision on detection.}
    \vspace{-5pt}
    \begin{center}
    \small
    \begin{tabular}{c|c|c|c|c|c}
    \hline
    & \multicolumn{4}{c|}{Dense RepPoints} &  \multirow{2}{*}{Mask R-CNN}\\
    \cline{2-5}
    & n=9 & n=25 & n=49 & n=81 & \\
    \hline
    \hline
    w.o. Inst & 37.9 & 37.7 & 37.5 & 37.5 & 36.4 \\
    \hline
    w. Inst & 38.1 & 38.7 & 39.2 & 39.4 & 37.3 \\
    \hline
    improve & +0.2 & +1.0 & +1.7 & \textbf{+1.9} & +0.9 \\
    \hline
    \end{tabular}
    \end{center}
    \label{tab::multitask-points}
    \vspace{-15pt}
\end{table}

\vspace{-10pt}
\paragraph{Benefit of Dense RepPoints on detection.}
Instance segmentation benefits object detection via multi-task learning as reported in Mask R-CNN~\cite{Mask-rcnn}. Here, we examine whether \textit{Dense RepPoints} can improve object detection performance as well. Results are shown in Table~\ref{tab::multitask-points}. Surprisingly, \textit{Dense RepPoints} not only takes advantage of instance segmentation to strengthen object detection, but also brings greater improvement when more points are used. Specifically, when $n=81$, \textit{Dense RepPoints} improves detection mAP by $1.9$ points. As a comparison, Mask R-CNN improves by $0.9$ points compared to Faster R-CNN. This indicates that multi-task learning benefits more from better representation. This suggests that \emph{Dense RepPoints} models a finer geometric representation. This novel application of explicit multi-task learning also verifies the necessity of using a denser point set, and it demonstrates the effectiveness of our unified representation.

\begin{table*}
    \caption{Performance of instance segmentation on COCO~\cite{MSCOCO} \texttt{test-dev}. Our method significantly surpasses all other state-of-the-arts. '*' indicates training without ATSS~\cite{zhang2019bridging} assigner and 'jitter' indicates using scale-jitter during training.}
	\begin{center}
    \resizebox{0.8\linewidth}{!}{
	\begin{tabular}{ll@{\ \ }cccccccccc}
    \toprule
    Method & Backbone & epochs & jitter & $AP$ & $AP_{50}$ & $AP_{75}$ & $AP_S$ & $AP_M$ & $AP_L$\\
    \midrule
	Mask R-CNN~\cite{Mask-rcnn}    & ResNet-101    & 12  &            & 35.7 &  58.0 & 37.8 & 15.5 & 38.1 & 52.4 \\
	Mask R-CNN~\cite{Mask-rcnn}    & ResNeXt-101&  12  &            & 37.1 & 60.0 & 39.4 & 16.9 & 39.9 & 53.5 \\
	TensorMask~\cite{TensorMask}   & ResNet-101    &  72  & \checkmark & 37.1 & 59.3 &  39.4 & 17.4 & 39.1 & 51.6 \\
    SOLO~\cite{wang2019solo}       & ResNet-101    &  72  & \checkmark & 37.8 & 59.5 & 40.4 & 16.4 & 40.6 & 54.2 \\
    \midrule
	ExtremeNet~\cite{ExtremeNet}   & HG-104     &  100 & \checkmark & 18.9 & - & - & 10.4 & 20.4 & 28.3 \\
	PolarMask \cite{PolarMask}     & ResNet-101 &  24  & \checkmark & 32.1 &  53.7 & 33.1 & 14.7 & 33.8 & 45.3 \\
    \midrule
    Ours*                           & ResNet-50    &  12  &            & 33.9 & 55.3 & 36.0 & 17.5  & 37.1 & 44.6 \\
    Ours                           & ResNet-50    &  12  &            & 34.1 & 56.0 & 36.1 & 17.7  & 36.6 & 44.9 \\
    Ours                           & ResNet-50    &  36  & \checkmark & 37.6 & 60.4 & 40.2 & 20.9 & 40.5 & 48.6   \\
    Ours                           & ResNet-101    &  12  &            & 35.8 & 58.2 & 38.0 & 18.7  & 38.8 & 47.1 \\
    Ours                           & ResNet-101    &  36  & \checkmark  & 39.1 & 62.2 & 42.1 & 21.8  & 42.5 & 50.8 \\
    Ours                           & ResNeXt-101&  36  & \checkmark & 40.2 &  63.8 & 43.1 & 23.1 & 43.6 & 52.0 \\
    Ours                           & ResNeXt-101-DCN & 36 &
    \checkmark & 41.8 & 65.7 & 45.0 & 24.0 & 45.2 & 54.6 \\
    \bottomrule
	\end{tabular}
	}
	\end{center}

	\label{tab::inst_system}
    \vspace{-8pt}
\end{table*}

\subsection{Comparison with other SOTA methods}
A comparison is conducted with other state-of-the-arts methods in object detection and instance segmentation on the COCO \texttt{test-dev} set. We use 729 representative points by default, and trained by distance transform sampling and set-to-set supervision. ATSS~\cite{zhang2019bridging} is used as the label assignment strategy if not specified. In the inference stage, the instance mask is generated by adopting triangulation as post-processing.

We first compare with other state-of-the-art instance segmentation methods. Results are shown in Table~\ref{tab::inst_system}. With the same ResNet-101 backbone, our method achieves $39.1$ mAP with the 1x setting, outperforming all other methods. By further integrating ResNeXt-101-DCN as a stronger backbone, our method reaches $41.8$ mAP.

\begin{table*}
\caption{Object detection on COCO~\cite{MSCOCO} \texttt{test-dev}. Our method significantly surpasses all other state-of-the-arts. '*' indicates training without ATSS~\cite{zhang2019bridging} assigner and 'jitter' indicates using scale-jitter during training.}
\begin{center}
\resizebox{0.8\textwidth}{!}{
\begin{tabular}{ll@{\ \ }ccccccccc}
\toprule
Method & Backbone & epochs & jitter & $AP$ & $AP_{50}$ & $AP_{75}$ & $AP_{S}$ & $AP_{M}$ & $AP_{L}$ \\
\midrule
Faster R-CNN\cite{FPN}     & ResNet-101     & 12 &              & 36.2 & 59.1 & 39.0 & 18.2 & 39.0 & 48.2     \\
Mask R-CNN\cite{Mask-rcnn} & ResNet-101     & 12  & &    38.2 & 60.3 & 41.7 & 20.1 & 41.1 & 50.2    \\
Mask R-CNN\cite{Mask-rcnn} & ResNeXt-101 & 12 &  & 39.8 & 62.3 & 43.4 &  22.1 &  43.2 &  51.2    \\
RetinaNet\cite{RetinaNet}  & ResNet-101     & 12 & &  39.1 &59.1 & 42.3 & 21.8 & 42.7 & 50.2    \\
RepPoints\cite{RepPoints}  & ResNet-101     & 12 &  & 41.0 & 62.9 & 44.3 & 23.6 & 44.1 & 51.7  \\
ATSS\cite{zhang2019bridging} & ResNeXt-101-DCN & 24 & \checkmark   & 47.7  &66.5 & 51.9 & 29.7 & 50.8 &  59.4 \\
\midrule
CornerNet\cite{CornerNet}    & HG-104       & 100 & \checkmark  &  40.5 & 56.5 & 43.1 & 19.4 & 42.7 & 53.9 \\
ExtremeNet\cite{ExtremeNet}  & HG-104       & 100 & \checkmark  &  40.1 & 55.3 & 43.2 & 20.3 & 43.2 & 53.1 \\
CenterNet~\cite{CenterNet}   & HG-104       & 100 & \checkmark  &  42.1 & 61.1 & 45.9 & 24.1 & 45.5 & 52.8 \\
\midrule
Ours*                        & ResNet-50       & 12 &             &  39.4   & 58.9  &  42.6 & 22.2 & 43.0 & 49.6  \\
Ours                        & ResNet-50       & 12 &             &  40.1   & 59.7 & 43.3  & 22.8 & 42.8 & 50.4  \\
Ours                        & ResNet-50       & 36 & \checkmark &  43.9 & 64.0 & 47.6 & 26.7 & 46.7 & 54.1  \\
Ours                        & ResNet-101       & 12 &             &  42.1    & 62.0 & 45.6  & 24.0 & 45.1 & 52.9  \\
Ours                        & ResNet-101       & 36 & \checkmark  &  45.6 & 65.7 & 49.7  & 27.7 & 48.9 & 56.6  \\
Ours                        & ResNeXt-101   & 36 & \checkmark  &  47.0 & 67.3 & 51.1  & 29.3 & 50.1 & 58.0 \\
Ours                        & ResNeXt-101+DCN   & 36 & \checkmark  &  48.9 & 69.2 & 53.4  & 30.5 & 51.9 & 61.2 \\
\bottomrule
\end{tabular}
}
\end{center}

\label{table::det_system}
\vspace{-10pt}
\end{table*}
We then compare with other state-of-the-arts object detection methods. Results are shown in Table~\ref{table::det_system}. With ResNet-101 as the backbone, our method achieves $42.1$ mAP with the 1x setting, outperforming RepPoints~\cite{RepPoints} and Mask R-CNN by $1.1$ mAP and $3.9$ mAP, respectively. With ResNeXt-101-DCN as a stronger backbone, our method achieves $48.9$ mAP, surpassing all other anchor-free SOTA methods, such as ATSS~\cite{zhang2019bridging} which only obtains $47.7$ mAP.

\subsection{Upper Bound Analysis}
\label{sec::empirical}
We design two oracle experiments to reveal the full potential of our method. 

\begin{figure}
\begin{center}
\includegraphics[width=0.5\textwidth]{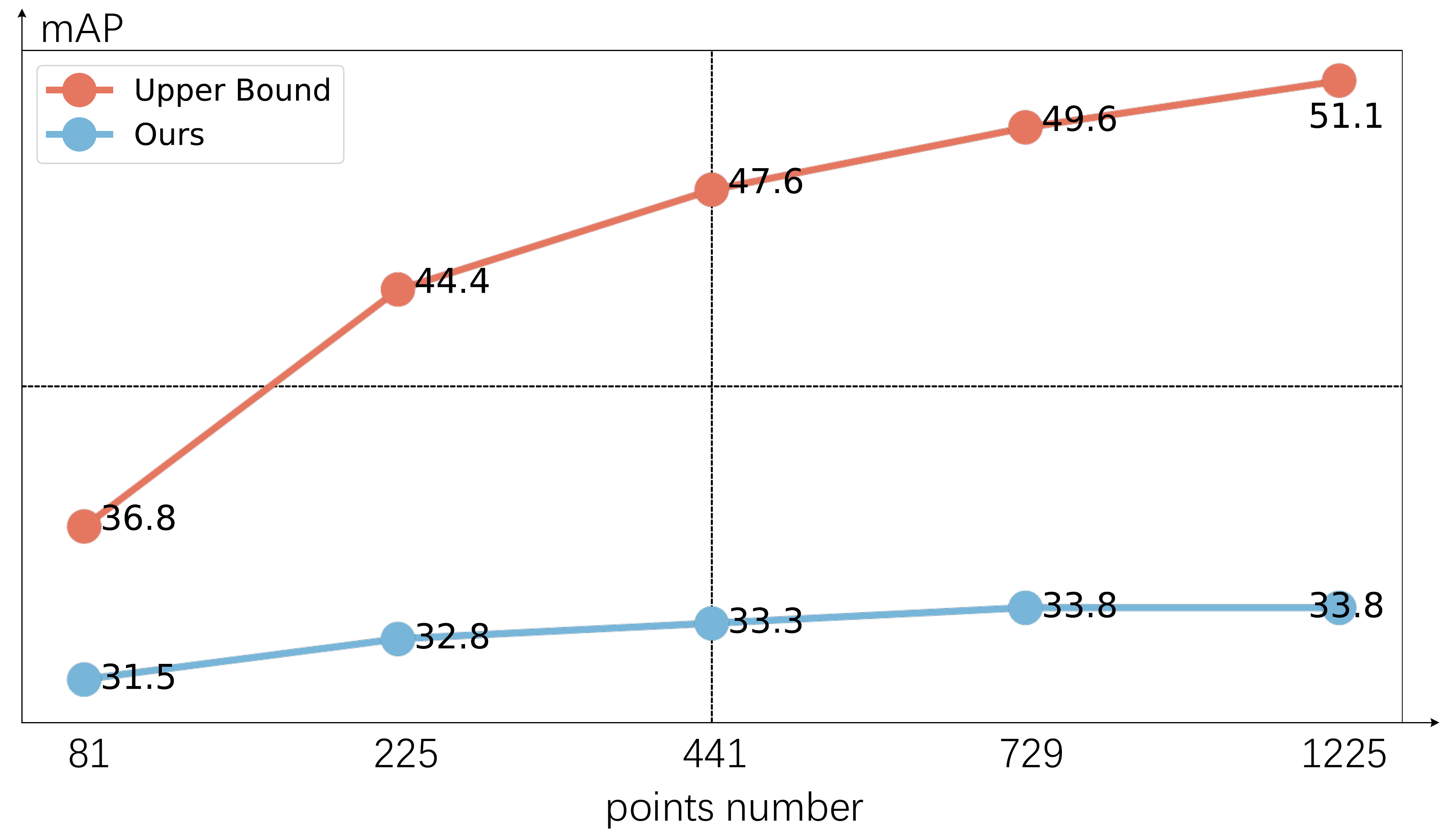}
\end{center}
\caption{Illustration for upper bound of \textit{Dense RepPoints}.}
\label{fig::upper}
\vspace{-15pt}
\end{figure}

\vspace{-10pt}
\paragraph{Upper bound of attribute scores.} The first experiment shows how much gain can be obtained when all the learned attribute scores are accurate and the learned point locations remain the same. In this experiment, we first calculate the IoU between predicted bboxes and ground-truth bboxes to select positive samples (IoU threshold=$0.5$). Then, we change the predicted attribute scores of these positive samples to ground-truth scores. The attribute scores of negative samples remain the same. Finally, we utilize these new attribute scores to generate binary masks. This experiment on a ResNet-50 backbone yields about 39.4 detection mAP (the fluctuation of detection performance under different numbers of points is negligible). Results are shown in Figure \ref{fig::upper}. We observe large performance gain when the attribute scores are absolutely accurate, which suggests that our method still has great potential if the learned attribute scores are improved. When the number of points increases to $1225$, the upper bound performance can improve nearly $60\%$ over the original segmentation performance. Clearly, a better detection result (better point locations) will also boost the upper bound of our mask representation. 

\begin{table}[ht]
\caption{The upper bound of IoU between predicted masks and ground-truth using DTS and triangulation post-processing under different point numbers.}
\begin{center}
\begin{tabular}{c|c|c|c|c|c|c|c}
\hline
n & 9 & 25 & 49 & 81 & 225 & 441 & 729 \\
\hline
IoU & 53.9 & 70.2 & 78.5 & 84.3 & 91.2 & 94.3 & \textbf{95.6} \\
\hline
\end{tabular}
\end{center}
\label{tab::iou}
\vspace{-30pt}
\end{table}

\noindent\textbf{Upper bound of DTS and triangulation}
The second experiment examines the upper bound when all the attribute scores and point locations are equal to the ground truth. First, we use DTS to generate points for each ground-truth mask. Then we assign ground-truth attribute scores to these points. Finally, we use triangulation interpolation to predict masks. Table \ref{tab::iou} shows the average IoU of our predicted masks and ground-truth masks. It can be seen that the IoU is nearly perfect (above $95\%$) when the points number increases, which indicates that our DTS and triangulation post-processing method can precisely depict the mask.

\section{Conclusion}

In this paper, we present \emph{Dense RepPoints}, a dense attributed point set representation for 2D objects. By introducing efficient feature extraction and employing dense supervision, this work takes a step towards learning a unified representation for top-down object recognition pipelines, enabling explicit modeling between different visual entities, $e.g.$ coarse bounding boxes and fine instance masks.
Besides, we also propose a new point sampling method to describe masks, shown to be effective in our experiments.
Experimental results show that this new dense 2D representation is not only applicable for predicting dense masks, but also can help improve other tasks such as object detection via its novel multi-granular object representation.
We also analyze the upper bound for our representation and plan to explore better score head designs and system-level performance improvements particularly on large objects.

\vspace{1.0em}
\noindent\textbf{Acknowledgement} We thank Jifeng Dai and Bolei Zhou for discussion and comments about this work. Jifeng Dai was involved in early discussions of the work and gave up authorship after he joined another company.

{\small
\bibliographystyle{ieee_fullname}
\bibliography{dense-pts}
}

\end{document}